\documentclass{article}
\usepackage{spconf,amsmath,graphicx}
\usepackage{microtype}
\usepackage{graphicx}
\usepackage{subfigure}
\usepackage{booktabs} 
\usepackage{multirow}
\usepackage{enumitem}
\usepackage{bbold}
\usepackage{stfloats}
\usepackage{hyperref}
\usepackage[capitalize,noabbrev]{cleveref}
\usepackage{soul,color}
\newcommand{\cmark}{$\surd$}%
\newcommand{\xmark}{$\times$}%
\newcommand{\model}{VLCap }

\title{VLCap: Vision-Language with Contrastive Learning \\ for Coherent Video Paragraph Captioning}
\name{Kashu Yamazaki, Sang Truong, Khoa Vo, Michael Kidd, Chase Rainwater, Khoa Luu, Ngan Le\thanks{This material is based upon work supported in part by the US National Science Foundation, under Award No. OIA-1946391, NSF 1920920.}}
\address{University of Arkansas, Fayetteville, AR 72701 USA}
\begin{document}
%
\maketitle
\begin{abstract}
\vspace{-1mm}
In this paper, we leverage the human perceiving process, that involves vision and language interaction, to generate a coherent paragraph description of untrimmed videos. We propose vision-language (VL) features consisting of two modalities, i.e., (i) vision modality to capture global visual content of the entire scene and (ii) language modality to extract scene elements description of both human and non-human objects (e.g. animals, vehicles, etc), visual and non-visual elements (e.g. relations, activities, etc). Furthermore, we propose to train our proposed VLCap under a contrastive learning VL loss. The experiments and ablation studies on ActivityNet Captions and YouCookII datasets show that our VLCap outperforms existing SOTA methods on both accuracy and diversity metrics. Source code: \url{https://github.com/UARK-AICV/VLCAP}

\end{abstract}
\begin{keywords}
Contrastive Learning, Video Captioning, Vision, Language
\end{keywords}
\vspace{-5mm}
\section{Introduction}
\vspace{-2mm}
\label{sec:intro}
Video paragraph captioning (VPC) aims to generate a paragraph description of untrimmed videos with several temporal event locations in a coherent storytelling. VPC can be considered as a simplified version of dense video captioning by eliminating the requirements for generating event proposals. VPC takes a video with its corresponding event proposals as the input and returns a coherent paragraph as the output. A typical VPC contains two components corresponding to (i) feature extraction to encode each event into a feature and (ii) caption generation to decode features into a list of sentences. An essential requirement of VPC is maintaining the intra-event coherence between words within a sentence describing an event and inter-event coherence between sentences within a paragraph describing an entire video.

\begin{figure}[!t]
    \centering
    \includegraphics[width=0.9\linewidth]{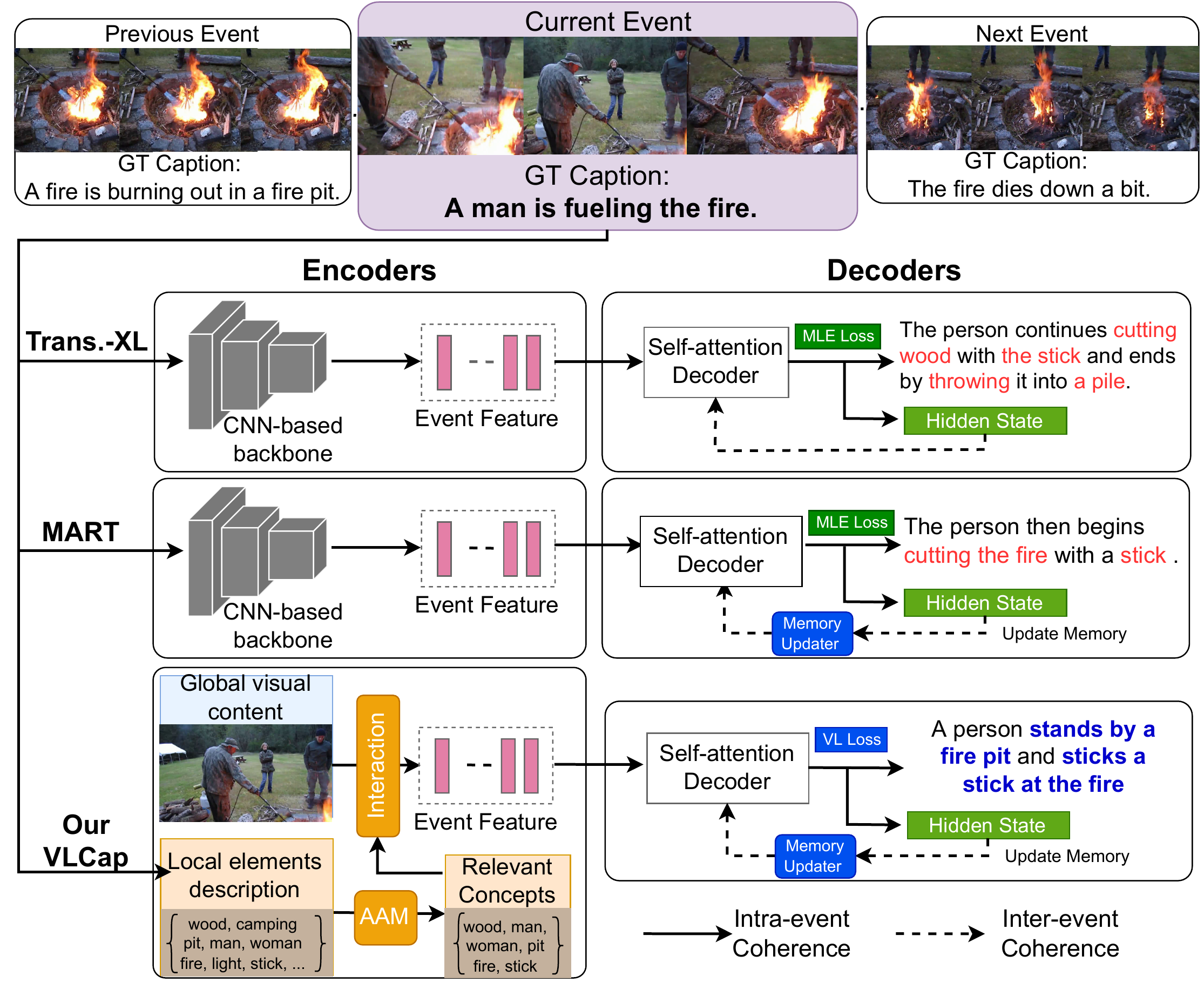}
    \vspace{-3mm}
    \caption{A comparison between our proposed \model and recent SOTA VPC methods, e.g., Transformer-XL (Trans.-XL) \cite{dai2019transformer} and MART \cite{lei2020mart}. At the encoder: both Transformer-XL and MART encode visual features by applying a CNN-based backbone network whereas our \model encodes VL feature by an adaptive attention mechanism (AAM) \cite{vo2021aei} with two modalities, i.e. (i) global visual content, (ii) local elements description. At the decoder: we propose to utilize Transformer to model intra-event coherence and GRU memory to model inter-event coherence. While both  Transformer-XL and MART are trained by MLE loss, our VLCap is trained by our proposed VL loss. \vspace{-5mm}
    }
    \label{fig:compare_model}
\end{figure}

Zhou, et al. \cite{zhou2018end} first leveraged the success of Transformer \cite{vaswani2017attention} to dress VPC task, known as Vanilla Transformer VPC. In their approach, intra-event coherence is decoded by a Transformer but there is no mechanism to model the inter-event coherence i.e., each event is decoded individually. Later, \cite{xiong2018move} tackled this limitation and proposed MFT by utilizing LSTM \cite{hochreiter1997long}. In MFT, the last hidden state of the current sentence is used as an initial hidden state for the next sentence. However, the coherence between sentences in MFT is ill-favored, facing the gradient vanishing problem \cite{pascanu2013difficulty} and unable to model long-term dependencies \cite{hochreiter2001gradient}. Being inspired by the recent transformer language model, Transformer-XL \cite{dai2019transformer}, which is able to resolve context fragmentation for language modeling, \cite{lei2020mart} proposed MART. While Transformer-XL directly uses hidden states from previous segments, MART is designed as a unified encoder-decoder to prevent overfitting and reduce memory usage. 

\begin{figure*}[]
    \centering
    \includegraphics[width=0.75\linewidth]{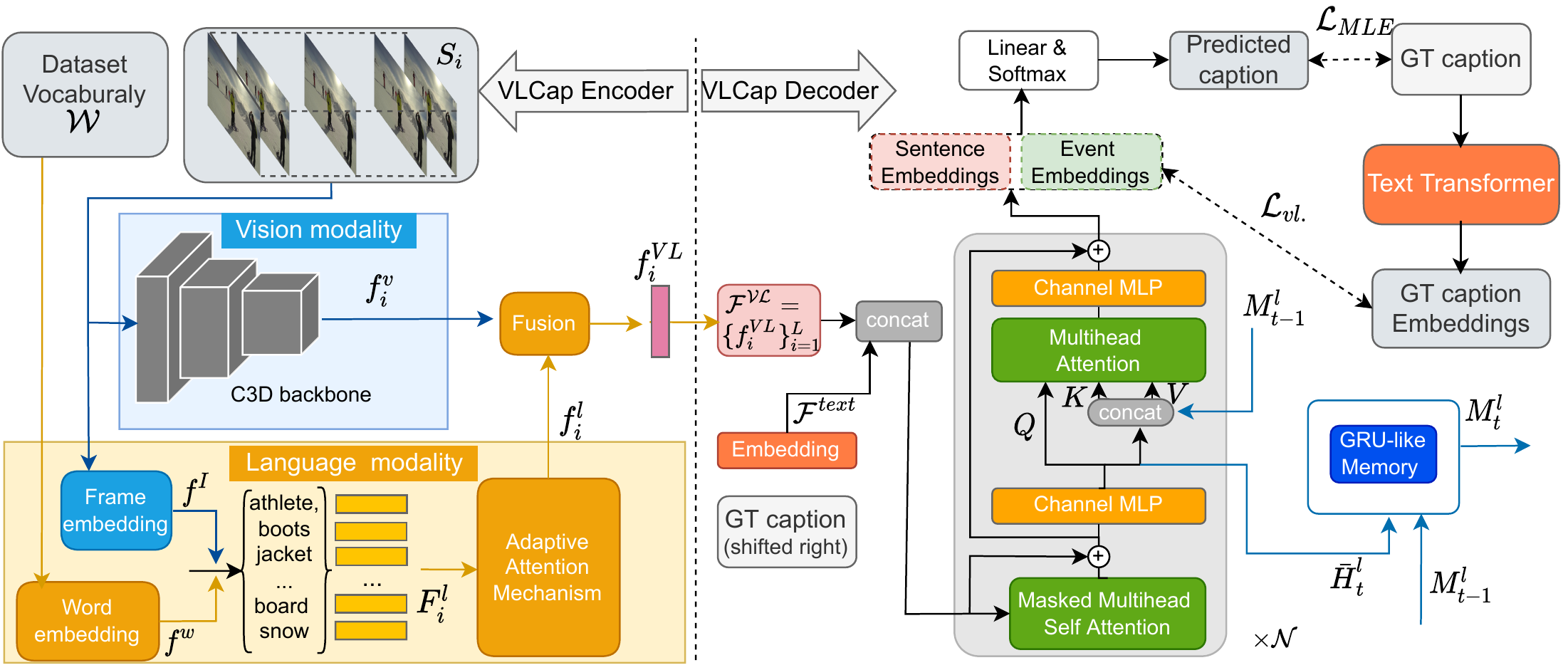}
    \vspace{-4mm}
    \caption{ Overall network architecture of our proposed \model consisting of two modules i.e. (i) VLCap Encoder (left) takes a snippet $S_i$ as an input and returns VL feature $f_i^{VL}$ as its output. VLCap Decoder (right) takes a list of VL features $\{f_i^{VL}\}_{i=1}^{L}$ extracted from $L$ snippets as its input and returns a predicted caption, which is then compared to the groundtruth caption by our proposed VL loss $\mathcal{L}_{VL} = \mathcal{L}_{MLE} + \mathcal{L}_{vl}$. \vspace{-5mm}}
    \label{fig:overall}
\end{figure*}

Clearly, to understand and describe a video, we not only observe the entire scene but also pay attention to both element scenes such as human and non-human objects (e.g., vehicles, animals, tools, etc.), visual and non-visual elements (e.g., actions, relations, etc). Furthermore, vision and language are two primary capabilities of humans language influences basic perceptual processing \cite{Lupyan2020}. However, most of the existing VPC approaches \cite{zhou2018end, park2019adversarial, dai2019transformer, zhou2019grounded, lei2020mart, wang2021end} decode caption description by applying a backbone, e.g. C3D \cite{C3D}, I3D \cite{carreira2017quo}, 2Stream \cite{2_stream_1, 2_stream_2}, or Slowfast \cite{SlowFast} to extract global visual information of the entire scene. By doing that, they ignore the interaction between the entire scene and relevant elements as well as disregard the fact that language and perception are two central cognitive systems. 

In this paper, we propose a multi-modal VL representation consisting of the global visual feature of the entire scene and linguistics relevant scene elements. While maximum likelihood estimation (MLE) is the most widely used loss function for supervised learning VPC, it does not guarantee that the learnt latent features represent the groundtruth captions. In this paper, we leverage contrastive learning \cite{hjelm2019learning, tian2020contrastive} and propose VL Loss, which consists of two terms corresponding to captioning loss ($\mathcal{L}_{cap.}$) and a contrastive contextual loss ($\mathcal{L}_{vl}$). The network comparison between our proposed VLCap with other existing VPC networks is shown in Fig. \ref{fig:compare_model}.

\vspace{-3.8mm}
\section{Proposed Method}
\vspace{-3mm}
Our proposed VLCap is designed as a unified encoder-decoder architecture and contains two main modules, i.e., VLCap Encoder and VLCap Decoder. Both modules are trained in an end-to-end framework by our proposed VL loss function. The entire architecture of \model is shown in Fig. \ref{fig:overall}.

In this section, we first introduce all notations and VPC problem formulation as follows: Given an untrimmed video $\mathcal{V}=\{v_i\}_{i=1}^\mathcal{|V|}$, where $|\mathcal{V}|$ is the number of frames, and a list of its important events $\mathcal{E}=\{e_i=(e^s_i, e^e_i)\}_{i=1}^{|\mathcal{E}|}$, where $|\mathcal{E}|$ is the number of events within a video and event $e_i$ is defined by a pair of beginning and ending timestamps $(e^s_i, e^e_i)$. Our objective is to generate a coherent paragraph $\mathcal{P}=\{\textbf{s}_i\}_{i=1}^{|\mathcal{E}|}$ that describes the whole video $\mathcal{V}$. In this setup, a sentence $\textbf{s}_i$ aims to describe its corresponding event $e_i$. We use notation $e = (e^s, e^e)$ to denote an event and it is presented by a sequence of frames $\mathcal{V}_e = \{v_i|e^s\leq i \leq e^e\}$.
\vspace{-3.2mm}
\subsection{VLCap Encoder}
\vspace{-1.5mm}
This module aims to extract VL feature $\mathcal{F}^{VL}$ given an event $e$, presented by a sequence of frames $\mathcal{V}_e$. Follow the standard setup \cite{zhou2018end, park2019adversarial, dai2019transformer, zhou2019grounded, lei2020mart, wang2021end}, we divide  $\mathcal{V}_e$ into $L$ snippets, $\{S_i\}_{i=1}^L$, each snippet $S_i$ consists of $\delta$ consecutive frames, where $L=\bigr\lceil \frac{|\mathcal{V}_e|}{\delta} \bigr\rceil$ and $|\mathcal{V}_e|$ is the number frames in $\mathcal{V}_e$. VLCap Encoder processes a snippet $S_i$ to extract $f_i^{VL}$. As a result, VLCap Encoder processes the event $e$ to extract feature $\mathcal{F}^{VL} = \{f_i^{VL}\}_{i=1}^L$ as shown in the Fig.\ref{fig:overall} (left). The VLCap Encoder contains three modalities as follows:

\noindent
\textbf{i. Vision Modality}
This modality aims to extract visual content by applying a C3D network \cite{C3D} into snippet $S_i$. The output feature map of C3D network $\phi$ is processed by average pooling to reduce the entire spatial dimension followed by channel multilayer perceptron (MLP). As a result, each snippet $S_i$ is represented by a feature $f_i^v$.
\vspace{-1.5mm}
\begin{equation}
    f_i^v = average\_pool(\phi({S_i}))
\end{equation}

\begin{table}[!t]
\centering
\caption{Datasets information. $2^{nd}$ col.: number of training videos; $3^{rd}$ cols.: number of validation videos. $4^{st}$ col.: number of event segments for each video on average. }
\begin{tabular}{l|l|l|l}
\hline
 Dataset & train & val & \shortstack{\#event \\ / video}\\ \hline
 
\shortstack{ActivityNet Captions~\cite{krishna2017dense}} & 10,009                 & 4,917  & 3.65                                    \\ \hline
YouCookII~\cite{zhou2018towards}          & 1,333                  &  457        & 7.7 \\
\hline
\end{tabular}
\label{tab:dataset}
\vspace{-5.2mm}
\end{table}

\begin{table*}[!t]
\centering
\caption{Performance comparison of \model with other SOTA models on ActivityNet Captions \textit{ae-val}. $\dag$ denotes results by us.}
\resizebox{0.9\linewidth}{!}{
\begin{tabular}{l|l|l|cccc|cc}
\toprule
\label{tab:anet_val}
Methods & Year & Input & B@4 $\uparrow$ & M $\uparrow$ &  C $\uparrow$ & R $\uparrow$  & Div@2 $\uparrow$ & R@4 $\downarrow$ \\ \hline
Vanilla Transformer~\cite{zhou2018end} & CVPR2018 & Res200 + Flow & 9.75 & 15.64& 22.16 & 28.90$^\dag$ & \underline{77.40}$^\dag$ & 7.79 \\
AdvInf \cite{park2019adversarial} & CVPR2019 & C3D + Object & 10.04 & \underline{16.60} & 20.97 & -- & -- &  5.76 \\ 
GVD \cite{zhou2019grounded} & CVPR2019 & Res200 + Flow + Object &  11.04 & 15.71 &  21.95 & -- & -- & 8.76\\ 
Transformer-XL \cite{dai2019transformer} & ACL2019 & Res200 + Flow & 10.39 & 15.09 & 21.67 & 30.18$^\dag$ & 75.96$^\dag$ & 8.54 \\ 
Transformer-XLRG \cite{lei2020mart} & ACL2020 & Res200 + Flow & 10.17 & 14.77 & 20.40 & -- & -- & 8.85 \\
MART \cite{lei2020mart}  & ACL2020   & Res200 + Flow &  10.33 & 15.68 & 23.42 &  \underline{30.32}$^\dag$ & 75.71$^\dag$ & \underline{5.18} \\
PDVC \cite{wang2021end} & ICCV2021 & C3D + Flow & \underline{11.80} &  15.93 &  \underline{27.27}  & -- & -- & --\\ 
\hline
\textbf{\model} (ours) & -- & C3D + Language  & \textbf{14.00} & \textbf{17.78} & \textbf{32.58} & \textbf{36.37} & \textbf{78.01} & \textbf{4.42} \\
\bottomrule
\end{tabular}
}
\vspace{-4mm}
\end{table*}

\begin{table*}[!t]
\centering
\caption{Performance comparison of \model with other SOTA models on ActivityNet Captions \textit{ae-test}. $\dag$ denotes results by us.} 
\resizebox{0.9\linewidth}{!}{
\begin{tabular}{l|l|l|cccc|cc}
\toprule
Methods & Year & Input & B@4 $\uparrow$ & M $\uparrow$&  C $\uparrow$ & R $\uparrow$ & Div@2 $\uparrow$  & R@4 $\downarrow$\\ \hline
Vanilla Transformer \cite{zhou2018end}& CVPR2018 & Res200 + Flow & 9.31 & 15.54&  21.33& 28.98$^\dag$ &  \underline{77.29}$^\dag$ & 7.45\\
Transformer-XL \cite{dai2019transformer} & ACL2019 &  Res200 + Flow & 10.25 & 14.91 & 21.71 & 30.25$^\dag$ & 76.17$^\dag$ &8.79 \\ 
Transformer-XLRG \cite{lei2020mart} & ACL2020& Res200 + Flow & 10.07 & 14.58 & 20.34 & -- & --&  9.37 \\


MART \cite{lei2020mart} & ACL2020 & Res200 + Flow &  9.78 & 15.57 & 22.16 & \underline{30.85}$^\dag$ & 75.69$^\dag$ & \underline{5.44} \\
MART w/ COOT \cite{ging2020coot} & NIPS2020& COOT & \underline{10.85} &  \underline{15.99} & \underline{28.19} & -- & -- &  6.64 \\
\hline
\textbf{\model} (ours) & -- & C3D + Language & \textbf{13.38} & \textbf{17.48} & \textbf{30.29} & \textbf{35.99} & \textbf{78.29} &  \textbf{4.18}  \\
\bottomrule
\end{tabular}
}
\label{tab:anet_test}
\vspace{-4mm}
\end{table*}

\noindent
\textbf{ii. Language Modality} 
This modality aims to extract element-level linguistic details of each snippet $S_i$. We leverage the success of recent works \cite{Patashnik2021styleclip, Yang2021} which have proved the effectiveness of feature representation learned via Contrastive Language-Image Pre-training (CLIP) \cite{radford2021learning}. Given a snippet $S_i$, the linguistic feature $f_i^l$ is extracted by the following steps: (i) - Word embedding: We construct a vocabulary $\mathcal{W} = \{w_1, \dots w_N\}$ using the groundtruth captions from training dataset. Each word $w_i\in \mathcal{W}$ is encoded by a Transformer network \cite{vaswani2017attention} into a text feature $f_i^w$. We then project feature $f_i^w$ onto text projection matrix $W_t$ pre-trained by CLIP to obtain word embedding word i.e. $w^e =  W_t \cdot f^w$, where $f^w = \{f_i^w\}_{i=1}^{m}$. (ii) - Language-based frame embedding: We choose the middle frame $I$ to present each snippet $S_i$. We first encode frame $I$ by a pre-trained Vision Transformer \cite{dosovitskiy2020image} to extract visual feature $f^I$. We then project feature $f^I$ onto visual projection matrix $W_i$ pre-trained by CLIP to obtain image embedding $I^e =  W_i\cdot f^I$. The pairwise cosine similarity between embedded $I^e$ and $w^e$ is then computed. Top $k$ similarity scores are chosen as language-based frame embedding feature $F_i^l$. (iii) -  language feature extraction: In this step, we employ Adaptive Attention Mechanism (AAM) \cite{vo2021aei} to select the most relevant representative language features:
\vspace{-1.2mm}
\begin{equation}
f_i^{l} = \text{AAM}(F_i^l) = \text{AAM}(\text{cosine}(I^e, w^e))
\vspace{-1.2mm}
\end{equation}


\begin{table*}[!b]
\centering
\caption{Performance comparison of \model with other SOTA models on YouCookII validation set.}
\resizebox{0.9\linewidth}{!}{
\begin{tabular}[h]{l|l|l|cccc|cc}
\toprule
Methods & Year & Input & B@4 $\uparrow$ & M $\uparrow$&  C$\uparrow$ & R $\uparrow$  & Div@2 $\uparrow$ & R@4 $\downarrow$\\ \hline

Vanilla Transformer~\cite{zhou2018end} & CVPR2018 & Res200 + Flow & 4.38 & 11.55 & 38.00 & -- &-- & --\\
GPaS \cite{zhang2020dense} &  IEEE-TM2020 &  Res200 &  1.64 & 12.20 & 41.44 & \underline{27.98} & --& --  \\
MART \cite{lei2020mart}  & ACL2020  &  Res200 + Flow & 8.00   & 15.90 & 35.74 & -- &-- & \textbf{4.39} \\
MART w/ COOT \cite{ging2020coot} &  NIPS 2020 & COOT & \underline{9.44} & \textbf{18.17} & \underline{46.06} & -- &-- & 6.30\\
\hline
\textbf{\model} (ours) & -- & C3D + Language &  \textbf{9.56} &  \underline{17.95} &  \textbf{49.41}  & \textbf{35.17} & 67.97 & \underline{5.16}  \\
\bottomrule
\end{tabular}
}
\label{tab:youcook}
\vspace{-5mm}
\end{table*}

\noindent
\textbf{iii. Fused Modality}
This modality aims to fuse the visual feature $f_i^e$ and linguistic feature $f_i^{l}$ given a snippet $S_i$. We first extract the inter-feature relationships by utilizing a self-attention layer \cite{vaswani2017attention}. We then merge them by a mean operation:
\vspace{-1.3mm}
\begin{equation}
    f_i^{VL} = \text{mean}(\text{Self-Attention}([f^v_i;f^l_i ]))
\vspace{-1.2mm}
\end{equation}

\vspace{-3mm}
\subsection{VLCap Decoder}
\vspace{-3mm}
Leveraging the recent success of Transformer vision-language models \cite{Chen2019uniter, lei2020mart}, we adopt the unified encoder-decoder Transformer to generate a caption of a given event $e_i$ consisting of $L$ snippets, i.e., $e_i = \{S_i\}_{i=1}^L$. By applying VLCap Encoder, each $S_i$ is presented by a VL feature $f_i^{VL}$, thus the event $e_i$ is presented by $\mathcal{F}^{VL} = \{f_i^{VL}\}^{L}_{i=1}$. Let $\mathcal{F}^{text}$ denotes textual tokens. Both the video features $\mathcal{F}^{VL}$ and textual tokens $\mathcal{F}^{text}$ are taken as a unified input for the Transformer layers i.e., 
\vspace{-2mm}
\begin{equation} 
H_t^0 = [\mathcal{F}^{VL}, \mathcal{F}^{text}]
\end{equation}

\vspace{-1mm}
To model inter-event coherence, our unified encoder-decoder Transformer is equipped with GRU-like memory to remember history information. Inspired by MART \cite{lei2020mart}, at step time $t$, decoding the $t^{th}$ event, the $t^{th}$ layer aggregates the information from both its intermediate hidden states $\bar{H}^l_t$ and the memory states $M_{t-1}^l$ from the last step, using a multi-head attention. The input key, value, query matrices are $K, V = [M_{t-1}^l; \bar{H}^l_t]$, $Q = \bar{H}^l_t$. A feed forward layer is then used to encode the memory augmented hidden states. The output is then merged with $\bar{H}^l_t$ using a residual connection and layer norm to obtain the hidden states output ${H}^l_t$. The process is as follows:
\vspace{-1mm}
\begin{equation}
\begin{split}
  U^l_t & = \text{MultiHeadAtt}(M_{t-1}^l, \bar{H}^l_t, \bar{H}^l_t) \\
  R^l_t & = \text{tanh}(W_{mr}^lM_{t-1}^l + W_{ur}^lU_{t}^l + b^l_r) \\
  Z^l_t & = \text{sigmoid}(W_{mz}^lM_{t-1}^l + W_{uz}^lU_{t}^l + b^l_z) \\
  M^l_t & = (1 - Z^l_t) \odot    R^l_t + Z^l_t \odot    M^l_{t-1}
\end{split}
\end{equation}
where $\odot$ is Hadamard product and $W_{mr}, W_{ur}, W_{mz}, W_{uz}$ are network parameters and $b^l_r, b^l_z$ are bias.
\vspace{-3.5mm}
\subsection{VL Loss}
\vspace{-2mm}
Maximum likelihood estimation (MLE) loss, which is trained to increase the likelihood between predicted captions groundtruth, is the most common in VPC. However, it is unable to address the question of how well the learnt latent features represent the groundtruth captions. In this paper, we proposed Visual-Linguistic (VL) Loss, which tackles the aforementioned concerns while maintaining the likelihood between predicted caption groundtruth. Particularly, we leverage the recent advantages of constractive learning to propose $\mathcal{L}_{vl}$ to pull all snippets of the same event and push snippets of different events. Let consider a set of $N$ events $\{e_i\}_{i=1}^N$, each event $e_i$ consists of $L$ snippets $\{S_i\}_{i=1}^L$. Each event $e_i$ has its corresponding groundtruth caption $\textbf{c}$, which is then presented as $f^\mathcal{T}$ by the pretrained Text Transformer from CLIP \cite{radford2021learning}. Apply our proposed VLCap network into $e_i$, we obtain the event embeddings $\mathcal{F}_i$ which is then processed as a vector $f_i = \text{mean}(\mathcal{F}_i)$. $\mathcal{L}_{vl}$ is computed as follows:
\vspace{-2mm}
\begin{equation}
\footnotesize
    \mathcal{L}_{vl} = -\sum_{i,j=1}^{N} \mathbb{1}_{i=j}\log e^\rho( f_{i} \cdot f_{j}^\mathcal{T})
    +  \mathbb{1}_{i\neq j}(1-\log e^\rho( f_{i} \cdot f_{j}^\mathcal{T}))
\vspace{-2mm}
\end{equation}

where $\rho$ is a learnable temperature parameter, which is initialized to $\log(1/0.07)$, to prevent scaling of the dot product values and reduce training instability.

Our VL loss $\mathcal{L}_{VL}$ consists of two terms corresponding to caption-caption loss ($\mathcal{L}_{MLE}$) and a vision-language loss ($\mathcal{L}_{vl}$) as follows:

\begin{equation}
\mathcal{L}_{VL} = \mathcal{L}_{MLE} + \mathcal{L}_{vl}
\end{equation}

\vspace{-4mm}
\begin{table*}[!b]
\centering
\caption{Ablation study on the contribution of the proposed VL feature and VL loss $\mathcal{L}_{VL}$ on ActivityNet Captions dataset.}
\resizebox{\linewidth}{!}{
\begin{tabular}{l|llll|llllll|llllll}
\toprule
&     &      &     & & \multicolumn{6}{c|}{ae-test}  & \multicolumn{6}{c}{ae-val} \\ \cline{6-17}
 Exp. & Vision & Lang & $\mathcal{L}_{MLE}$ & $\mathcal{L}_{VL}$ & B@4$\uparrow$ & M$\uparrow$ & C$\uparrow$ & R$\uparrow$ & Div@2$\uparrow$ & R@4$\downarrow$ & B@4$\uparrow$ & M$\uparrow$ & C$\uparrow$ & R$\uparrow$ & Div@2$\uparrow$ & R@4$\downarrow$ \\ \hline
\#1 & \cmark & \xmark & \cmark & \xmark & 11.10 & 15.72 & 27.68 & 31.75 & 74.34 & 7.11 & 11.50 & 16.05 & 28.83 & 31.85 & 74.17 & 7.32 \\
\#2& \cmark & \xmark & \xmark & \cmark & 11.17 & 16.27 & \underline{30.22} & 31.72 & \textbf{79.18} & \textbf{3.54} & 11.57 & 16.40 & 29.92 & 31.87 & \textbf{79.00} & \textbf{3.85}\\
\#3& \cmark & \cmark & \cmark & \xmark & \underline{13.56} & \underline{17.42} & 30.10 & \underline{35.78} & {77.36} & {4.77} & \underline{13.59} & \underline{17.49} & \underline{30.80} & \underline{35.83} & {77.06} & {5.06} \\
\#4& \cmark & \cmark & \xmark & \cmark & \textbf{13.38} & \textbf{17.48} & \textbf{30.29} & \textbf{35.99} & \underline{78.29} &  \underline{4.18} &
\textbf{14.00} & \textbf{17.78} & \textbf{32.58} & \textbf{36.37} & \underline{78.01} & \underline{4.42}\\
\bottomrule
\end{tabular}
}
\label{tb:ablation}
\end{table*}

\section{Experiments}
\vspace{-2.3mm}
\subsection{Datasets, Metrics and Implementation Details}
\vspace{-2mm}

We benchmark our \model on two popular VPC datasets, YouCookII~\cite{zhou2018towards} and ActivityNet Captions~\cite{krishna2017dense}. Information of those datsets are summarized in Table \ref{tab:dataset}. We follow the previous works \cite{lei2020mart} to split the original validation set into two subsets: \textit{ae-val} with 2,460 videos for validation and \textit{ae-test} with 2,457 videos for test.

We benchmark VLCap on four standard accuracy metrics, i.e., BLEU@4 (B@4) \cite{papineni2002bleu}, METEOR (M) \cite{denkowski2014meteor}, CIDEr (C) \cite{vedantam2015cider}, ROUGE (R) \cite{lin2004rouge} and two diversity metrics i.e., 2-gram diversity (Div@2) \cite{div} and 4-gram repetition (R@4) \cite{xiong2018move}. 

Adam optimizer was used to train our \model with an initial learning rate of 1e-4, $\beta_1=0.9$, $\beta_2=0.999$, $L_2$ weight decay of 0.01, and learning rate warmup over the first 5 epochs. During the training, we use the label smoothing with a value of 0.1 and $\lambda=0.1$. 
\vspace{-4mm}

\subsection{Performance and Comparison}
\vspace{-2mm}

Tables \ref{tab:anet_val}, \ref{tab:anet_test} report the performance comparison on ActivityNet Captions corresponding to \textit{ae-val} and \textit{ae-test} sets whereas Table \ref{tab:youcook} shows the performance comparison on YouCookII validation set. In each table, we highlight the best and the second-best with \textbf{bold} and \underline{underline}. On YouCookII, VLCap obtains the best performance on B@4, C and R metrics whereas it gains compatible on other metrics. On ActivityNet Captions, VLCap obtains the best performance with large gaps on both accuracy metrics and diversity metrics compared to the second-best score. Take ActivityNet Captions as an example, corresponding to \textit{ae-val} and \textit{ae-test} sets, our VLCap gains (2.2\%/1.18\%/5.31\%/6.05\%) and (2.53\%/1.49\%/3.10\%/5.14\%) higher on BLEU@4/METEOR/CIDEr/ROUGE metrics while improves (0.61\%) and (1.0\%) on Div@2 as well as reduces (0.31\%) and (1.26\%) on R@4 compare to the second-best achievement.  

To evaluate the effectiveness of our proposed VL feature as well as VL loss, we conduct ablation studies as shown in Table \ref{tb:ablation}. The capability of the proposed VL loss ($\mathcal{L}_{VL}$) is shown in comparisons between Exp.\#1 v.s \#2 and  Exp.\#3 v.s \#4 where we compare between VL loss and MLE loss. The advantage of the proposed VL feature is shown in comparisons between Exp.\#1 v.s \#3 and Exp.\#2 v.s \#4 where we compare between vision feature (i.e. C3D) and VL feature. Both of our proposed VL feature and VL loss contribute in improving the accuracy and diversity metrics.

\vspace{-4mm}
\section{Conclusion}
\vspace{-1mm}
In this work, we present a novel VLCap network for video paragraph captioning. Our VLCap network is trained in an end-to-end framework with a two-fold contribution: (i) VL feature, which extracts the global visual features of the entire scene and local linguistics feature of scene elements; and (ii) VL loss, which is trained by a constrative learning mechanism. In VLCap network, the intra-event coherence is learnt by a Transformer whereas the inter-event coherence is modeled by GRU-like memory. Comprehensive experiments and ablation studies on ActivityNet Captions and YouCookII datasets demonstrate the effectiveness of our VLCap, which outperforms the existing SOTA approaches on both accuracy (BLEU@4, METEOR, CIDEr, ROUGE) and diversity (Div@2, R@4) metrics.




\small
\bibliographystyle{IEEEbib}
\bibliography{main}

\end{document}